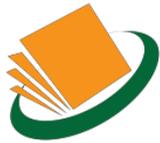

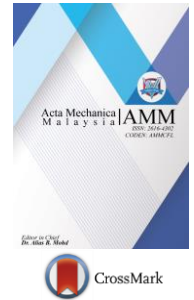

REVIEW ARTICLE

# STRATEGIC EVALUATION IN OPTIMIZING THE INTERNAL SUPPLY CHAIN USING TOPSIS: EVIDENCE IN A COIL-WINDING MACHINE MANUFACTURER

Dilip U Shenoy[1], Vinay Sharma[2], Shiva Prasad, H C [3*]

[1]Senior Officer, Technology Excellence Group, QuEsT Global, Bengaluru, Karnataka, India. ORCID:0000-0001-9588-9663
[2]UG Student, Dept. of ICE, MIT, Manipal Academy of Higher Education, Manipal, India, ORCID: 0000-0002-7600-9744
[3]Faculty, Dept. of Mechanical Engineering, School of Automobile, Mechanical, and Mechatronics, Faculty of Engineering, Manipal University Jaipur, Jaipur, Rajasthan, India, ORCID 0000-0002-1296-8970
*Corresponding Author Email: shiva.prasad@jaipur.manipal.edu; hcs.prasad@manipal.edu

| ARTICLE DETAILS | ABSTRACT |
|---|---|
| *Article History:*<br><br>Received 25 August 2019<br>Accepted 29 September 2019<br>Available online 01 January 2020 | Most of manufacturing firm aims to optimize their Supply Chain in-terms of improved profitability of their products through value Addition. This study takes a critical look into the factors that affect the Performance of internal supply chain with respect to specific criteria's. Accordingly, ranking these factors to get the critical dimensions of supply chain performance in the manufacturing industry. A semi-structured interview with the pre-defined set of questions used to collect the responses from decision makers of the firm. Multi criteria decision-making tool called TOPSIS is used to evaluate the responses and rank the factors. The results of this indicate that supplier relationship and inventory planning were most principal factors positively influencing on-time delivery of the product, production flexibility, cost savings, additional costs. This study helps to identify and optimize the process parameters using objective and subjective evaluation approach. The combined influence of thought process of manager to optimize the internal supply chain is extracted in this work.<br><br>**KEYWORDS**<br><br>Manufacturing, value stream map, cycle time, supply-chain, strategies, a technique for an order of preference by similarity to the ideal solution. |

## 1. INTRODUCTION

The Indian manufacturing industry is resolutely marching towards steady growth in the global competitive scenario. Due to the swift pace of globalization, the evolution of modern industrial technologies and economic liberalization, the manufacturing sector is facing intense competition. The coil winding industry (electric coils are used for drone motors, toys, micromotors, solenoids, transformers etc.,) is expected to be scalable business opportunities across the country as well as the globe. The growth in the applications industries like small house old appliances, large industrial power transmitting, aircraft circuits, and marine application would steer the growth of coil winding manufacturing industry. Resulting in a balanced approach from supply to demand process and the organization must have maintain their competitive and get higher market share. The Government's role in providing promising investments, infrastructure facilities is very crucial on one side and manufacturing firm's strategies to reduce cost, improving process efficiency, and productivity plays an equivalent role on another side [1]. Hence it is critical to identify the key performance indicators of the internal supply chain process. Most organizations have a wide range of interdependent pool of operations that form its supply chain. Current business operations no longer work in silos, but terms of business processes. Hence it is essential to prioritize the strategies required to optimally operate the business processes with a lean philosophy in mind. There are great alternatives to strategies to keep the manufacturing supply chain in a healthy condition. It becomes necessary to prioritize the strategies required to optimally operate the business processes with lean philosophy in mind [2]. It is imperative for the firm to prioritize the strategies that impacts the criteria like on-time delivery, Cost-effectiveness with use of modern, advanced management information system and software tools like gamification [3].

This study provides the firm a clear visualization of the key strategies that directly affect the performance of their internal value stream, to take a quick action on the key strategies first and realize the value addition to their supply chain. The list of strategies or factors that determine the performance are prioritized based on specific criteria relevant to Cost and Benefit Attributes using a multi-criteria decision-making tool called Technique for the order of preference by similarity to the ideal solution (TOPSIS). To understand the thought process of each of the Heads of Departments in the supply chain, with reference to value added and non-value added activities in their department, the responses were collected and analyzed using semi-structured interview approach. Time study and physical walk through to all the departments and interaction with the shop floor stakeholder, were carried out. After validating responses from interviews and data collected from physical walkthrough, six alternatives/factors affecting the internal supply chain and four criteria are selected for the study that are elaborated in the following sections. A score-sheet were given to each Head of Department to score each factor from a scale of one to nine to with respect to the criteria. Scores indicate the correlation between each factor and criteria. Score one being least correlation and nine being highest correlation. The criterions are weighted and Multi Criteria Decision Making (MCDM) technique enabled Technique for the Order of Preference by Similarity to Ideal Solution (TOPSIS) is used to prioritize the alternatives [4].

## 2. LITERATURE REVIEW

The effectiveness of a firm's supply chain is measured in several ways, and the factors used to measure effectiveness is specific to the way the business operates [5]. Every firm develops a measurement metric called Key performance indicator (KPIs) based on the firm's vision and mission. For some, delivering products on-time is a KPI, and for others, it is supplier reliability or sales volume. Several authors have identified several critical KPIs for best supply chain management process as total delivered cost, customer service, supplier variability, demand variability, operating costs, performance to plan and inventory [5,6]. It is very crucial to have a performance measurement system in supply chain management.

However, the measuring metrics should be strategically decided based on relevancy to the business operations. Morgan says the performance management system is essential to give managers a sense of handling out-of-control situations [7,8].

Validating the supply chain factors is essential to measure what performance indicators are relevant to the organization. Always there are a wide variety of alternatives for performance indicators. Many researchers have attempted to aggregate a conventional measuring instrument and variables for Supply chain performance and management effectiveness, as Gunasekaran and Kobu have consolidated the (i) Balanced scorecard perspective (ii) Components of performance measures (iii) Location of measures in supply chain links (iv) Decision making-levels (v) Nature of measures (vi) Measurement base (vii) Traditional vs. modern measures [6]. The manufacturing firm where this study was considered and the operational environment, some key performance measuring alternatives scales are chosen. Moreover, TOPSIS proves out to be one of the appropriate mathematical tools to decide the best among alternatives w.r.t set of attributes.

## 3. SUPPORTING WORKS ON TOPSIS

Technique for the order of preference by similarity to ideal solution (TOPSIS) was developed by Wang and Yoon in 1981 [9]. Past researchers have shown the application of TOPSIS to evaluate the alternatives in a manufacturing organization. According to Hwang and Yoon the essential benefit of multi-criteria decision-making tools are useful for evaluating both qualitative as well as quantitative attributes [9].

On the verge of getting the leading-edge in-service market, Choudhury has used the TOPSIS method to evaluate the performance of Higher Education Institutes in business management using customer perceived Service quality dimensions [10]. Sarkar has used AHP and TOPSIS to evaluate technologies on electrical energy options [9]. Gladysz and Santarek have utilized fuzzy-TOPSIS approach to assess the RFID technology for Logistics operation of the manufacturing company [11]. Kumar and Singh have used the combination of fuzzy AHP and TOPSIS to evaluate third-party logistics in a supply chain [12]. TOPSIS has also been utilized to evaluate plant location in the supply chain [13]. Based on short-term goals and long-term goals has used entropy weight and TOPSIS II to evaluate synchronized supply chain [14]. These kinds of literature and more proven that TOPSIS is the tool to utilize and evaluate a set of alternative w.r.t to specific criteria. This survey analyzes a set of six alternatives w.r.t four criteria's that indicates the key performance indicator of the supply chain.

## 4. METHODOLOGY

### 4.1 Overview of the firm's supply chain

The analysis was carried out in an ISO 9000:2008 certified coil winding machine manufacturing company XQZ2 in South India. They are the leading manufacturer and exporter of semi-automatic and automatic coil winding machines. The company designs the entire product (coil winding machine) and outsources the manufacturing of the parts to a list of vendors. The parts are assembled in-house and then packed and shipped and sold in the country as well as exported. The business supply chain of the firm has three levels; the first level is the strategic Business process with emphases on Business development. The second level is the Business process, from sales and manufacturing to meet the client's needs. The third is a SIPOC, supplier, incoming inspection and assembling as per customer (see Figure 1).

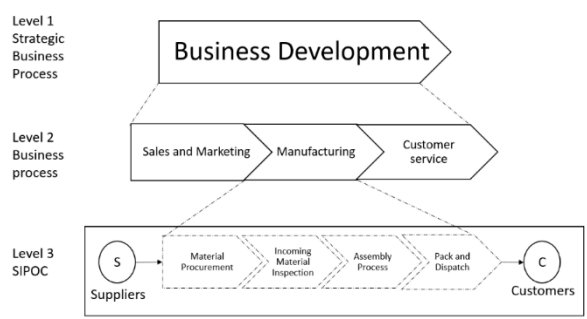

**Figure 1:** Business supply chain overview

### 4.2 Framework of study

This study is a blend of descriptive and applied research wherein a case is described, and action is applied into it to get some insights (see Figure 2).

A survey was conducted in the firm and responses were collected using a scoresheet, where each question measures a specific alternative w.r.t criteria set.

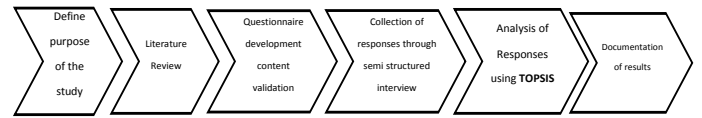

**Figure 2:** Conceptual framework

TOPSIS is a multi-criteria decision-making tool that gives the best alternative in a set of alternatives based on scoring the alternative in a suitable scale (Likert, seven points or nine-point) concerning a set of criteria. The central principle of TOPSIS lies in finding the closeness to positive ideal solution(an alternative that has the best scores). The analysis is purely based on the judgments given by the scorers. The framework of TOPSIS is as shown: (Figure 3).

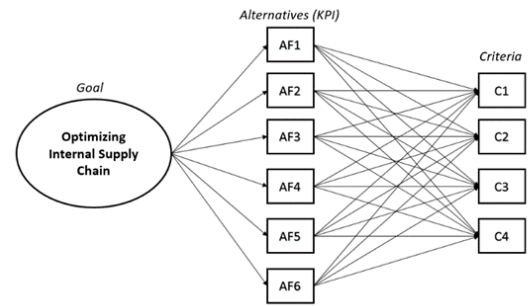

**Figure 3:** TOPSIS Framework

There are six alternative factors, and it is scored on a scale of nine in comparison with four criteria (See Figure 3).

*4.2.1 Selection of alternatives or factors and criteria*

The study place where the firms' operational firms environment, a set of factors were considered as the performance indicators by spending a month time at each department and interacting with the heads. Key responsibilities of each department were analyzed. Based on the discussion with the top management and the heads of various departments, six alternatives were given, and four criteria for these six alternatives were chosen. The alternatives include information sharing across the departments (AF1), Supplier relationship in-terms of supplier part quality, reliability, and variability (AF2), Information technology (AF3), Inventory Planning (AF4), 5S in the shop floor (AF5), Overall Labour effectiveness (AF6) and the criteria were on-time delivery of the product (C1), Production flexibility (C2), Cost-effectiveness (C3), Additional cost (C4). C1, C2, C3 were benefited attributes, and C4 was the cost attribute.

*4.2.2 Alternative Factors*

AF1) On-time information sharing: This factor measures whether sharing information on-time across different departments can influence the four criteria's namely On-time delivery of the product, production flexibility, cost savings, additional cost.

AF2) Supplier Relationship: Supplier relationship is one of the essential criteria to evaluate supply chain performance. The organizations that depend on suppliers for the parts that are not manufactured in-house must have a healthy relationship with the suppliers. Here supplier relationship includes all the dimensions like supplier reliability, lead time, variability, quality.

AF3) Information Technologies: With the modernization of industries, the Business pattern of the organization has changed. Departments no longer work in silos. There must be a centralized information system to efficiently function business operations. Hence this factor is used to check whether it has any influence on the criteria as mentioned earlier.

AF4) Inventory Planning: Inventory plays a crucial role in manufacturing organizations. To deliver products on-time, save costs the firm should economically order their inventories.

AF5) 5S in shop floor: 5S is a lean manufacturing concept developed by Japanese manufacturing giant Toyota. It indicates housekeeping on the shop floor — a place for everything and everything at its place.

AF6) Overall Labour Effectiveness (OLE): OLE is a measure of the effectiveness of the labor force in the organization. Technically OLE can be measured as the product of Availability, Quality, and Performance of the workforce.

The selected alternatives and criteria are as follows:
Alternative Factors:
$A_1$ : On-time information sharing
$A_2$ : Supplier Relationship
$A_3$ : Information Technology
$A_4$ : Inventory Planning
$A_5$ : 5S in the shop floor
$A_6$ : Overall Labour Effectiveness
Criteria:
$C_1$ : On-time delivery
$C_2$ : Production Flexibility
$C_3$ : Cost-effectiveness in-terms of cost savings
$C_4$ : Additional Cost

### 4.2.3 TOPSIS Calculations

STEP 1: Determine the objective of the study and relevant criteria. Here the objective of the study was to identify factors affecting the supply chain performance.

STEP 2: Construct the decision matrix based on the score. There are two types of attributes in criteria – Benefit attribute & Cost attribute. Ex: On-time delivery of a product is a benefit attribute; i.e. more score indicates better on-time delivery that are desirable. Whereas say the manufacturing cost of the product is a cost attribute, i.e. more score means more manufacturing cost that are undesirable. The decision matrix is a two-dimensional matrix where each row is dedicated to an alternative (The factor responsible for supply chain performance). Each column is dedicated to a criterion. So, the score $S_{ij}$ represents the impact of ith alternative w.r.t jth criteria (Table 1).

**Table 1:** Decision Matrix

| Decision matrix $D_{ij}$ | | Criteria | | | |
|---|---|---|---|---|---|
| | | $C_1$ | $C_2$ | $C_3$ | $C_4$ |
| Alternatives | $A_1$ | $S_{11}$ | $S_{12}$ | $S_{13}$ | $S_{14}$ |
| | $A_2$ | $S_{21}$ | $S_{22}$ | $S_{23}$ | $S_{24}$ |
| | $A_3$ | $S_{31}$ | $S_{32}$ | $S_{33}$ | $S_{34}$ |
| | $A_4$ | $S_{41}$ | $S_{42}$ | $S_{43}$ | $S_{44}$ |
| | $A_5$ | $S_{51}$ | $S_{52}$ | $S_{53}$ | $S_{54}$ |
| | $A_6$ | $S_{61}$ | $S_{62}$ | $S_{63}$ | $S_{64}$ |

$S_{ij}$ – Score for $i^{th}$ alternative w.r.t $j^{th}$ criteria.

Table 2 shows the decision matrix obtained based on the responses.

**Table 2:** Decision matrix $D_{ij}$ based on the response
$w_{ij}$– assigned weights for each criteria

STEP 3: Obtain the normalized decision matrix $N_{ij}$. The decision matrix

| Decision matrix $D_{ij}$ | | Criteria | | | |
|---|---|---|---|---|---|
| | | $C_1$ | $C_2$ | $C_3$ | $C_4$ |
| Alternatives | $A_1$ | 7 | 6 | 7 | 7 |
| | $A_2$ | 8 | 8 | 7 | 6 |
| | $A_3$ | 7 | 6 | 6 | 6 |
| | $A_4$ | 8 | 7 | 8 | 6 |
| | $A_5$ | 6 | 6 | 6 | 6 |
| | $A_6$ | 7 | 8 | 6 | 6 |
| $w_{ij}$ | | 0.5 | 0.1 | 0.3 | 0.1 |

obtained in step 2 is normalized, to get normalized scores. The formula used to convert each score (S) to normalize score (R) is obtained by (1)

$$R = \frac{S}{\sqrt[2]{\sum_{j=1}^{6} S^2}} \quad (1)$$

R= Normalized Score
S = Score from Table 1

STEP 4: Obtain the weighted normalized decision matrix. Multiply each normalized score (R) by the assigned weight ($w_{ij}$) equation (2). So weighted normalized score with the weighted normalized decision matrix (Table3).

$V_{ij} = R_{ij}*w_{ij}$ (2)

**Table 3:** Weighted normalized decision matrix

| Weighted Normalized decision matrix | | Criteria | | | |
|---|---|---|---|---|---|
| | | $C_1$ | $C_2$ | $C_3$ | $C_4$ |
| Alternatives | $A_1$ | 0.1985 ($v_{11}$) | 0.0356 ($v_{12}$) | 0.1279 ($v_{13}$) | 0.0463 ($v_{14}$) |
| | $A_2$ | 0.2269 ($v_{21}$) | 0.0474 ($v_{22}$) | 0.1279 ($v_{23}$) | 0.0397 ($v_{24}$) |
| | $A_3$ | 0.1985 ($v_{31}$) | 0.0356 ($v_{32}$) | 0.1096 ($v_{33}$) | 0.0397 ($v_{34}$) |
| | $A_4$ | 0.2269 ($v_{41}$) | 0.0415 ($v_{42}$) | 0.1461 ($v_{43}$) | 0.0397 ($v_{44}$) |
| | $A_5$ | 0.1702 ($v_{51}$) | 0.0356 ($v_{52}$) | 0.1096 ($v_{53}$) | 0.0397 ($v_{54}$) |
| | $A_6$ | 0.1985 ($v_{61}$) | 0.0474 ($v_{62}$) | 0.1096 ($v_{63}$) | 0.0397 ($v_{64}$) |

**Highest value in each column** — **Lowest value in each column**

STEP 5: Obtain the Positive ideal solution (PIS) and negative ideal solution (NIS). PIS is an ideal set containing the best score for each criterion. Moreover, NIS is an ideal set containing the least scores from each criterion.

PIS A+ = (0.2269 (v21), 0.0474 (v22), 0.1461 (v43), 0.0463 (v14))

NIS A- = (0.1702 (v51), 0.0356 (v52), 0.1096 (v53), 0.0397 (v24))

STEP 6: Calculate relative closeness to the ideal solution. In each column, find the distance of each score from the respective positive ideal score (Table 4). Similarly, find the distance of each score from the negative ideal score (Table 5).

**Table 4:** Relative distance of each score from its respective PIS score

| Relative distance score from PIS | | Criteria | | | | $S_i^+ = \sqrt[2]{\sum_{i=1}^{4} vi^2}$ |
|---|---|---|---|---|---|---|
| | | $C_1$ | $C_2$ | $C_3$ | $C_4$ | |
| Alternatives | $A_1$ | (0.2269-0.1985)² | (0.0474-0.0356)² | (0.1461-0.1279)² | (0.0463-0.0463)² | 0.001 |
| | $A_2$ | (0.2269-0.2269)² | (0.0474-0.0474)² | (0.1461-0.1279)² | (0.0463-0.0397)² | 0.0001 |
| | $A_3$ | (0.2269-0.1985)² | (0.0474-0.0356)² | (0.1461-0.1096)² | (0.0463-0.0397)² | 0.0014 |
| | $A_4$ | (0.2269-0.2269)² | (0.0474-0.0415)² | (0.1461-0.1461)² | (0.0463-0.0397)² | 0.0005 |
| | $A_5$ | (0.2269-0.1702)² | (0.0474-0.0356)² | (0.1461-0.1096)² | (0.0463-0.0397)² | 0.0038 |
| | $A_6$ | (0.2269-0.1985)² | (0.0474-0.0474)² | (0.1461-0.1096)² | (0.0463-0.0397)² | 0.0012 |

**Table 5:** Relative distance of each score from its respective NIS score

| Relative distance score from PIS | | Criteria | | | | $S_i^- = \sqrt[2]{\sum_{i=1}^{4} vi^2}$ |
|---|---|---|---|---|---|---|
| | | $C_1$ | $C_2$ | $C_3$ | $C_4$ | |
| Alternatives | $A_1$ | (0.1702-0.1985)² | (0.0356-0.0356)² | (0.1096-0.1279)² | (0.0397-0.0463)² | 0.000944 |
| | $A_2$ | (0.1702-0.2269)² | (0.0356-0.0474)² | (0.1096-0.1279)² | (0.0397-0.0397)² | 0.003259 |
| | $A_3$ | (0.1702-0.1985)² | (0.0356-0.0356)² | (0.1096-0.1096)² | (0.0397-0.0397)² | 0.001321 |
| | $A_4$ | (0.1702-0.2269)² | (0.0356-0.0415)² | (0.1096-0.1461)² | (0.0397-0.0397)² | 0.003628 |
| | $A_5$ | (0.1702-0.1702)² | (0.0356-0.0356)² | (0.1096-0.1096)² | (0.0397-0.0397)² | 0.000517 |
| | $A_6$ | (0.1702-0.1985)² | (0.0356-0.0474)² | (0.1096-0.1096)² | (0.0397-0.0397)² | 0.001181 |

STEP 7: Sort the closeness value in the descending order according to the closeness ratio (C) given equation (3),

$$C_i = \frac{S^-}{S^+ + S^-} \quad (3)$$

The $C_i$ value is calculated from the equation and the values are tabulated (Table 6). Rank the alternatives accordingly i.e. higher the closeness value more is the priority.

**Table 6:** Closeness Ratio

| Alternative | Ci | Rank |
|---|---|---|
| A1 | 0.492824578 | 4 |
| A2 | 0.850943262 | 1 |
| A3 | 0.266585065 | 6 |
| A4 | 0.729255514 | 2 |
| A5 | 0.269528495 | 5 |
| A6 | 0.497990187 | 3 |

## 5. RESULTS

It is observed that Alternative 2 i.e. Supplier relationship has highest Ci value indicating it is closer to the positive ideal solution. Hence the top priority is given to the supplier relationship. Then followed by Alternative 4 i.e. Inventory planning. Hence based on the employee perception Supplier relationship and Inventory planning is ranked the top factors that affects the on-time delivery of the product and save costs to the company. Table 7 shows the Ranking of factors:

**Table 7:** Ranking of factors effecting supply chain performance

| Rank | Factors |
| --- | --- |
| 1 | Supplier Relationship |
| 2 | Inventory Planning |
| 3 | On-time information Sharing |
| 4 | OLE |
| 5 | 5S |
| 6 | Information Technologies (ERP) |

## 6. DISCUSSION

The significance of the results obtained from implementing Lean principles using perceiptional information gather from different sources. The study addresses critical insights to research question rised can internal supply chain management process be optimized if yes the multi criteria's are involved in selecting the process. The alternatives such as On-time information sharing, Supplier Relationship, Information Technology, Inventory Planning, 5S in the shop floor, Overall Labour Effectiveness having criteria as on-time delivery, flexible production systems, cost-effectiveness, reducing unncessary cost are few indicators of performance evalaution in the internal supply chain [15]. This study gives insights on current status of the internal supply chain to identify the value added and non-value added activities that affects the performance of production process. It is observed that by having an efficent flow of information and material the supply chain runs effectively. Only material flow without relevant information leads to defects in quality. Hence material and information flow are correlated and like two sides of same coin. To measure the health of supply chain it is important to identify the key symptoms. Since the analysis involved finding out of certain factors effecting the supply chain is validated using multi criteria decision making tool. MCDM is an efficient tool to capture influence of certain factors effecting the performance and the technique TOPSIS is used investigate the influence of factors effecting supply chain in the perspective of the top decision makers of the firm. The observational aspects, primary data colelcetd abd the results obtained are compared back with the employees' opinion and found to be true. There is a scope for partcipative investigation of factors with taking employees into confidence and further dig at micro-level.

## 7. CONCLUSION

Performance indicators is an important process to validate the process and its analysis. Since the analysis involved finding out of certain factors effecting the supply chain it was validated using Multi criteria decision making tool. MCDM is an efficient tool to capture influence of certain factors effecting the performance. The study gave a crtical look into the chief factors affecting Supply chain performance based on the responses collected from Heads of Departments through interview and questionnaire and physical walk through to the shopfloor. The responses were validated using MCDM technique TOPSIS. Based on the results from TOPSIS it is observed that Supplier Relationship in-terms of On-time delivery, right-first-time Quality parts and an efficient Inventory planning for the critical parts indicated best performance for the internal supply chain. Hence having a reliable supplier base, an efficient inventory management practice, a clear production plan and resource utilization can significantly optimize the supply chain process.

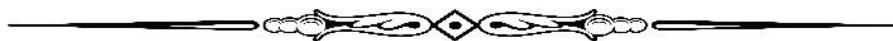